\pdfoutput=1

\documentclass[11pt]{article}

\usepackage{EMNLP2023}

\usepackage{times}
\usepackage{latexsym}

\usepackage[T1]{fontenc}

\usepackage[utf8]{inputenc}

\usepackage{microtype}

\usepackage{inconsolata}

\usepackage{booktabs}
\usepackage{graphicx}
\usepackage{multirow}
\usepackage{amsmath}
\usepackage{amssymb}
\usepackage{enumitem}
\usepackage{hyperref}
\usepackage[nameinlink]{cleveref}
\usepackage{listings} 

\usepackage{verbatim}

\usepackage[T1]{fontenc}

\usepackage[utf8]{inputenc}

\usepackage{microtype}

\usepackage{svg}
\usepackage{xcolor}
\usepackage{booktabs}
\usepackage{multirow}
\usepackage{float}
\usepackage{enumitem}
\usepackage{longtable}
\usepackage{supertabular}

\usepackage{dsfont}
\usepackage{amsmath}

\usepackage{todonotes}

\definecolor{codegreen}{rgb}{0,0.6,0}
\definecolor{codegray}{rgb}{0.5,0.5,0.5}
\definecolor{codepurple}{rgb}{0.58,0,0.82}
\definecolor{backcolour}{rgb}{0.95,0.95,0.92}

\lstdefinestyle{mystyle}{
    backgroundcolor=\color{backcolour},   
    commentstyle=\color{codegreen},
    keywordstyle=\color{magenta},
    numberstyle=\tiny\color{codegray},
    stringstyle=\color{codepurple},
    basicstyle=\ttfamily\small,
    breakatwhitespace=false,         
    breaklines=true,                 
    captionpos=b,                    
    keepspaces=true,                 
    numbers=left,                    
    numbersep=5pt,                  
    showspaces=false,                
    showstringspaces=false,
    showtabs=false,                  
    tabsize=2
}

\title{From RAG to RICHES: \underline{R}etrieval \underline{I}nterla\underline{ce}d with \underline{S}equence Generation}

\author{Palak Jain \qquad Livio Baldini Soares \qquad Tom Kwiatkowski \\
\qquad Google Deepmind  \\
\texttt{\{palakj,liviobs,tomkwiat\}@google.com}}

\newcommand{\loci}{\textsc{Riches}}

\newcommand{\llmm}{\textsc{PaLM2-M}}
\newcommand{\llml}{\textsc{PaLM2-L}}
\newcommand{\mydecomp}{\textit{Iter}}
\newcommand{\startsearch}{\texttt{<<}}
\newcommand{\stopsearch}{\texttt{>>}}

\begin{document}
\maketitle
\begin{abstract}
We present \loci{}, a novel approach that interleaves retrieval with sequence generation tasks. \loci{} offers an alternative to conventional RAG systems by eliminating the need for separate retriever and generator. It retrieves documents by directly decoding their contents, constrained on the corpus. Unifying retrieval with generation allows us to adapt to diverse new tasks via prompting alone. \loci{} can work with any Instruction-tuned model, without additional training. It provides attributed evidence, supports multi-hop retrievals and interleaves thoughts to plan on what to retrieve next, all within a single decoding pass of the LLM. We demonstrate the strong performance of \loci{} across ODQA tasks including attributed and multi-hop QA.


\end{abstract}

\section{Introduction}
Large language models (LLMs) have increasingly become the backbone for much of natural language processing and there has been a push to formulate a wide range of tasks as sequence to sequence transduction.
However, when LLMs need to interact with non-parametric knowledge in the form of an external evidence corpus, the typical approaches chain LLM generations with calls to a separate retrieval model as part of a multi-system pipeline.
In this paper we introduce a new approach, \loci{} (\underline{R}etrieval \underline{I}nterla\underline{ce}d with \underline{S}equence Generation) which can natively interleave text generations with retrievals from an evidence corpus using a single LLM and decoding process.

\begin{figure}[t]
\includegraphics[width=\linewidth]{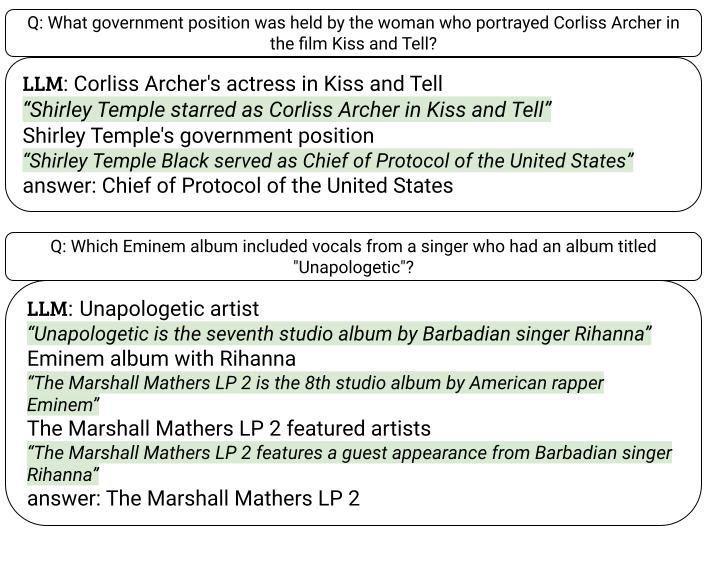}
\caption{Example \loci{} outputs for multi-hop queries with a single LLM and decoding pass. The green quoted text is "retrieved" or generated verbatim from the retrieval corpus. \loci{} generation natively interleaves thoughts and multiple retrieval evidences.}
\label{fig:intro}
\end{figure}

\loci{} builds on previous work that demonstrated the application of \emph{constrained decoding} to retrieval over a corpus \cite{jain20231,bevilacqua2022autoregressive} but extends this work to support multiple retrievals, entwined in a standard text generation procedure.
In this approach, we retrieve documents by directly decoding their contents or related natural language \emph{retrieval keys} that point to the documents they were generated from.
For example, Figure~\ref{fig:intro} illustrates a solution from \loci{} to multi-hop question answering \cite{yang2018hotpotqa}, where evidence must be retrieved from multiple separate documents, by iteratively generating an unconstrained `thought' about what needs to be retrieved and then generating a supporting proposition derived from an evidence corpus and tied to an original piece of supporting text. \loci{} executes this task in a single decoder pass.
For this example task, which is evaluated alongside others in Section~\ref{sec:results}, we have built on recent advances in chain-of-thought reasoning via prompting alone \cite{yao2022react} but have directly integrated the retrieval step without needing to account for any interaction with an external retrieval system.

The observations we build this work on are:
\begin{enumerate}
  \item \textit{LLMs are knowledge warehouses}: They internalise and generalise over vast quantities of training data and are often able to generate surprisingly accurate knowledge in response to complex inputs \cite{sun2022recitation}. However they are also susceptible to \emph{hallucination} and cannot account for fresh knowledge, not available at the time of training. That is where retrieval shines.
  \item \textit{LLM  decoding is a search process}: Language model decoders search for a single sequence in the set of all possible token sequences \citep{graves2012sequence}. Retrievers just need to constrain this search space to those sequences that are known to exist in a corpus of interest.
  \item \textit{Unifying tasks unlocks rapid development via prompting} By unifying retrieval with generation in a single decoder pass, we create a system that can be adapted to diverse new tasks via prompting alone, directly benefiting from the advances in instruction following. We later show that \loci{} works with an off-the-shelf instruction-tuned model, without any additional training. This is in contrast to pipelines that need to be rebuilt/retrained on a task-by-task basis.
\end{enumerate}

There is an another advantage of using language models as search agents. Of the two core operations in retrieval, indexing and search, indexing is constrained by corpus size, while search typically depends only on the index structure. Using large language models for indexing billion-token corpora is highly expensive, but search does not face the same bottle-neck. This enables us to unlock the knowledge stored in very large models for retrieval.

This work overlaps with a variety of related work focusing on retrieval, retrieval augmented generation \cite{lewis2020retrieval}, reasoning in language models, and open domain question answering.
We discuss their connections to \loci{} in Section~\ref{sec:related_work}, then introduce the key components of the generalizable \loci{} approach in Section~\ref{sec:approach}.

While \loci{} is applicable to any task that can be reduced to an interleaved generation of unconstrained text and pre-defined retrieval keys, we validate the approach with tasks in open domain question answering and show how it natively supports single-hop question answering, including the case where attribution to a source text is required; multi-hop question answering; and interleaving retrieval with `planning steps' that enhance the retrieval performance. Results are presented in Section~\ref{sec:results:overall} along with qualitative examples and analysis in Section~\ref{sec:results:qualitative} to help motivate the approach.

\section{Related Work}
\label{sec:related_work}

\paragraph{Retrieval Augmented Generation (RAG)}
ODQA tasks predominantly employ the RAG approach \cite{lewis2020retrieval} where typically a dense retriever \citep{karpukhin2020dense} retrieves documents from an evidence corpus and feeds to a language model for the final answer.
These pipelines involve switching between heterogeneous models and are hard to train in concert.
Moreover, Dense retrievers fail to generalize out-of-domain~\citep{thakur2021beir}.

\paragraph{Generative Retrieval}
\citep{metzler2021rethinking} techniques  shifting the onus of Search from non-parametric nearest neighbor scan to language models. Differentiable Search Index \citep{tay2022transformer} memorizes a mapping of query to opaque document identifiers, however memorization struggles to generalize to unseen  corpus \citep{pradeep2023does}.
An alternative approach is to use natural language keys as document identifiers, where keys are constrained decoded to lie in the corpus \citep{de2020autoregressive,bevilacqua2022autoregressive}. These systems still need an external model to generate answers.
1-Pager~\citep{jain20231} unifies evidence and answer generation, by generating a sequence of keywords that map to a document. However, isolated keywords limit context understanding and suffer similar pitfalls as lexical matching.

\paragraph{Recitation}
Separate from retrieval augmentation, language models have been shown to recite entire passages from memory \citep{sun2022recitation,yu2022generate}. But these passages are prone to hallucination. 
Our aim is to intersect contextual passage generation with corpus grounding.
GopherCite \citep{menick2022teaching}, a noteworthy work in this direction, generates quotes verbatim from a small set of documents using constrained decoding.
\loci{} aims to scale this to a billion-token corpus.

\paragraph{Iterative reasoning and Search}
In recent times, there have been several efforts to improve multi-hop question answering by better reasoning \citep{asai2023self} and planning \citep{adolphs2021boosting,yao2022react}. Language models have also been applied to the task of search to explore alternative paths \cite{yao2023tree,hao2023reasoning}.

Our work builds on these advances in reasoning while integrating search within generation.


\section{Retrieving while Generating}
\label{sec:approach}
We present a method of interleaving unconstrained text generation with the generation of \emph{retrieval keys} that point into a retrieval corpus.
For example, Figure~\ref{fig:intro} shows generations that interleave unconstrained `thoughts' with evidence sentences drawn from a predefined corpus for a multi-hop question answering task.
Later in this section we'll introduce a number of different choices of retrieval key as well as a variety of tasks that benefit from interleaved generation and retrieval.
However, for now we simply define a retrieval key as a sequence of tokens that exists in a pre-defined finite set of sequences $K$ where every entry is associated with one or more documents in an underlying corpus $C$.

Formally, we focus on the sequence to sequence transduction task where we predict an output sequence $\mathbf{y} = [y_0, \dots, y_n]$ conditioned on an input sequence $\mathbf{x} = [x_0, \dots, x_m]$ and we mark the start and end of a retrieval key in $y$ with special markers \startsearch{} and \stopsearch{}.
If we let $Q(\mathbf{y})$ be a function that returns all retrieval key spans from $y$ (i.e. $(i, j) \in Q([y_0, \dots, \startsearch{}, y_i, \dots, y_j \stopsearch{}, \dots, y_n])$) then we can update the standard autoregressive language modeling probability

\begin{align}
P_{\theta}(\mathbf{y}|\mathbf{x}) = \prod_{i=0}^{|\mathbf{y}|} P(y_i|y_0,\dots,y_{i-1}, \mathbf{x}, \theta)
\label{eqn:lm}
\end{align}
to include the indicator function $\mathds{1}_{K}(\mathbf{q})$ that maps elements of $K$ onto one and otherwise to zero.
\begin{align}
\begin{split}
P_{\theta}(\mathbf{y}|, \mathbf{x}, K) & = \frac{1}{Z}\prod_{\mathbf{q}\in \mathcal{Q}(\mathbf{y})} \mathds{1}_{\mathcal{K}}(\mathbf{q}) \\
& \times \prod_{i=0}^n P(y_i|y_0,\dots,y_{i-1}, \mathbf{x}, \theta)
\end{split}
\label{eqn:constlm}
\end{align}

where $Z$ is a normalizing term that accounts for the probability mass assigned by Equation~\ref{eqn:lm} to disallowed sequences.
In practice, we do not need to compute $Z$ and can sample from Equation~\ref{eqn:constlm} in the usual way, one token at a time, by simply zeroing out the probability of disallowed continuations as presented in Section~\ref{sec:constrained}.



\subsection{Constrained Beam Decoding}\label{sec:constrained}
\begin{figure*}[t]
\includegraphics[width=\textwidth]{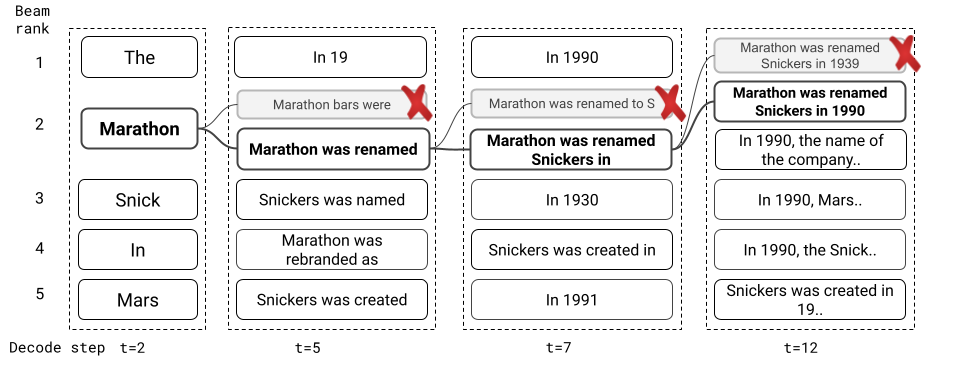}
\caption{Visualization of constrained beam for query: \textit{"when did marathon change its name to snickers?"}. The final \loci{} output is \textit{"Marathon was renamed Snickers in 1990"}. Bold boxes track the progress of the top-beam sequence. Grey crossed out boxes are sequences that the LLM preferred, but were blocked by corpus constraints.}
\label{fig:beam}
\end{figure*}
We opt for Beam Search \citep{graves2012sequence} as our decoding strategy to simulate a heuristic Best-first search. Here, the action or next node space is the entire vocab. At each time step, the LLM estimates the value of each node (token) given the paths explored so far and adds them to the fixed-size queue (Beam). \Cref{fig:beam} visualizes how the beam progresses over decoding timesteps. 
Unlike regular beam decoding where the top decoded sequences have only small variations, constraints impose sparsity over the search space resulting in diverse beams.
In \Cref{sec:adaptive}, we discuss how beam can hurt unconstrained generation and suggest hybrid decoding strategy as workarounds.
Constrained decoding can also gain from more sophisticated algorithms such as value-based decoding \citep{ren2017deep}, look-ahead scoring and planning \citep{lu2021neurologic, hao2023reasoning}.

\subsection{Efficient Constraints via the FM-Index}\label{sec:fm_index}
During decoding, model outputs are constrained to the corpus by masking out any continuation not in the corpus. To compute the continuations of a sequence, we use FM-index \citep{ferragina2000opportunistic}, a compressed suffix array augmented with additional data structures to support fast substring search operations. Unlike a Trie structure, it is also highly space economical due to the compression. 
Given a prefix, FM-Index can efficiently compute the next allowed tokens in O(Vocab), independent of the corpus-size. Below is the pseudo code for the modified decoding process.


\begin{lstlisting}[language=Python, label=code:constrained]
def constrain(input_prefix):
  # Fetch continuations for prefix
  allowed_tokens = fm_index.get_continuations(input_prefix)
  # Get next token probabilities
  logprobs = LLM.logprobs(input_prefix)
  # Disallowed tokens are set to -inf
  for i in logprobs:
    token = vocab[i]
    if token not in allowed_tokens:
        logprobs[i] -= np.inf
  return logprobs
\end{lstlisting}

\subsection{Adaptive Beam Size}\label{sec:adaptive}
In Section~\ref{sec:exp:evaluation} we introduce some tasks that interleave constrained and unconstrained generation.
The constrained generations must be precise---to match the target retrieval key exactly.
The unconstrained generations are generally more robust to small variations in surface form---these only need to convey the correct information to a reader, or to provide the model room for a `thought' trace when reasoning about a response.

To ensure that \loci{} can properly make use of beam search, which is here intended to ensure the model does not get stuck irretrievably after generating an incorrect constrained prefix, we introduce an adaptive decoding strategy that switches between full beam decoding for the sensitive constrained sequences but opts for greedy decoding when unconstrained. In practise, a constrained prefix is expanded to next beam-size probable tokens while an unconstrained prefix is expanded to only the next one token.
This is expected to provide room for rest of the beam to be utilized largely for constrained sequences. \Cref{sec:results:blend} shows experiments with multiple decode modes.

\subsection{Indexing Strategies}\label{sec:indexing}
The FM-Index used by \loci{} supports efficient indexing of all sub-strings in a corpus, which is useful when we want to generate corpus text verbatim.
However, it is not clear that this is the best option of retrieval key for the auto-regressive decoder in Section~\ref{sec:constrained}. A key question in index construction is the \textit{document representation} used in indexing.
In traditional lexical-based retrieval systems, documents are represented by the terms in it, with transformations such as stemming, weighing by corpus statistics \cite{robertson2009probabilistic}.
Neural retrieval systems transform raw text into dense vector representations and offload representation computation to the neural network. But even in this case, proper document chunking and/or multi-vector document significantly impact final performance~\cite{lee2021phrase,khattab2020colbert}.

In this section, we introduce a few different choices of retrieval keys, including a \emph{propositional index} that requires indexing time neural computation. A key consideration here is the interplay between the retrieval index and the search strategy.

\paragraph{Document Title and Section Headers} Many retrieval corpora such as Wikipedia have consistent structures in the form of titles and sometimes sub-titles and metadata. This provides a hierarchical structure such that one can first decode titles, sub-titles and then the document.

\paragraph{Paragraph Sub-string} A natural option for retrieval key is any sub-string of the unit of text being indexed itself. 
In most open domain question answering approaches, paragraph is the de-facto unit of evidence.
We can index paragraphs efficiently using the FM-index (Section~\ref{sec:fm_index}) and decode sub-strings directly with \loci{} to get pointers into the retrieval corpus.
It should be noted that this yields an inherently many-to-many mapping between paragraphs and retrieval keys, but that the mapping is in-effect one-to-one for longer sequences of tokens.

\paragraph{Sentence Sub-string} Similarly, individual sentences form a natural retrieval key. Sentence are smaller units of information than passage, but may not be interpretable or stand-alone.

\paragraph{Propositional Index} The above choices do not perform any non-trivial indexing step, unlike standard approaches in information retrieval where documents are mapped to sparse or dense vectors.
The omission of this indexing step may be desirable but it also forces \loci{} to deal with the non-uniform and diffused information in raw text.
An alternative that is closer, in intent, to the offline indexing step used by other IR systems, is to map each indexed chunk to a set of uniformly structured propositions \citep{min2023factscore,chen2022propsegment}. A proposition is a stand-alone unit that efficiently encodes small, atomic chunks of factual information. For example, instead of the sentence "He has 7M followers on Twitter" a proposition would be decontextualized to "Tom Cruise has 7M followers on Twitter."
We adopt a pre-existing propositional index from \citealt{chen2023dense} described in \Cref{sec:exp:datasets}. 

\Cref{sec:results:retrieval_keys} compares various Retrieval keys for the ODQA task with illustrations in \Cref{app:index_units}.

\section{Interleaving Retrieval and Generation}\label{sec:interleaving}
We have presented a method of interleaving unconstrained text generation with constrained generation of retrieval keys.
In this section we introduce a handful of tasks that make use of this interleaving either as a core task requirement, or as a means to an end by interleaving `thoughts' with retrieval actions to help guide search.
\paragraph{Attributed Question Answering} We apply \loci{} to the open domain question answering (ODQA) task where we score both the ability to correctly predict a short answer string and retrieve attribution for that answer \cite{bohnet2022attributed}.
See \Cref{tab:wins} for examples.
\paragraph{Multi-hop Question Answering} Interleaving between generation and retrieval can be powerful in multi-hop reasoning, where the model needs to retrieve and stitch together knowledge from multiple sources.
Examples of \loci{} outputs for multi-hop QA are given in \Cref{tab:win_multihop}.

\paragraph{"Thinking" for Retrieval}
Multi-step questions often require breaking down a query into smaller steps and reasoning or planning what to retrieve next. Foreshadowing retrieval with thoughts is crucial in this context. It helps direct the retrieval process, avoid repetitions, and, more importantly, allows for iterating upon and correcting previously erroneous retrievals. A few such demonstrations can be found in \Cref{tab:win_multihop}.

\begin{table*}[ht]
\begin{center}
\small
\begin{tabular}{p{0.25\textwidth}|p{0.3\textwidth}|p{0.35\textwidth}}
\toprule
\textbf{Query} & \textbf{Dense Retriever (GTR-xxl)} & \textbf{\loci{} Retrieval Output}  \\
\midrule
\multicolumn{3}{c}{\texttt{\# Leveraging Parametric knowledge}} \\
\midrule
who was the actor that played ben stone on law and order &
Benjamin Stone was a former Law and Order character & Ben Stone was played by Michael Moriarty \\
\midrule
jharkhand which festival is associated with cattle worship &
Gaijatra is the festival of cows
 &  Sohrai is a cattle festival of Jharkhand. \\
\midrule
\multicolumn{3}{c}{\texttt{\# Better Context understanding}} \\
\midrule
who plays at the prudential center in newark &
The Prudential Center is in Newark.
 & The Prudential Center is home to the New Jersey Devils \\
\midrule
when was coffee first made into a drink &
In 1724, society in Lima discovered an exotic drink called coffee.
 &  The earliest credible evidence of coffee-drinking or knowledge of the coffee tree appears in the middle of the 15th century. \\
\bottomrule
\end{tabular}

\caption{Example of \loci{} vs Dense Retrieval for single-hop QA. Only the retrieved text is shown for illustration.}
\label{tab:wins}

\end{center}
\end{table*}

\begin{table*}[ht]
\begin{center}
\small
\begin{tabular}{p{0.3\textwidth}|p{0.7\textwidth}}
\toprule
 \textbf{Query} & \textbf{\loci{} Output}  \\
 \midrule
 who had a baby at 100 in the bible
 \newline
 \newline
 (\# \texttt{Keywords guide retrieval})
 &  keyword: Sarah's age at birth of Isaac
\newline 
<< Sarah was 90 years old when Isaac was born. >> 
\newline
keyword: Abraham's age at birth of Isaac 
\newline
<< Abraham was 100 years old when Isaac was born. >>
\newline
answer: Sarah and Abraham \\
\midrule
when did the united states start using the death penalty
 \newline
 \newline
 (\# \texttt{keywords help recover from inaccurate retrieval})
 & 
keyword: Capital punishment in the United States
\newline
<< Capital punishment was reinstated in the United States in 1976. >> 
\newline
keyword: First execution in the United States 
\newline
<< The first recorded execution in the United States took place in 1608 at the Jamestown Colony in Virginia. >>
\newline
answer: 1608 \\
\midrule
What distinction is held by the former NBA player who was a member of the Charlotte Hornets during their 1992-93 season and was head coach for the WNBA team Charlotte Sting?
\newline
(\texttt{\# Planning what to retrieve})
& keyword: member of the Charlotte Hornets during their 1992-93 season 
\newline
<< Muggsy Bogues played for the Charlotte Hornets from 1988 to 1997. >>
\newline
keyword: Muggsy Bogues' distinction
\newline
<< Muggsy Bogues is the shortest player ever to play in the National Basketball Association (NBA). >>
\newline
answer: shortest player ever to play in the National Basketball Association (NBA) \\
\bottomrule
\end{tabular}

\caption{Example Iterative retrieval outputs from \loci{}. Remarks are annotated as \texttt{(\# Comments)}}
\label{tab:win_multihop}

\end{center}
\end{table*}


\section{Experimental Setup}

\subsection{Datasets}
\label{sec:exp:datasets}
\paragraph{Queryset} Our experiments are focused on open domain question answering tasks including both single and multi-hop benchmarks. For single-hop, we use the Open-NQ \cite{kwiatkowski2019natural} dataset. To evaluate multi-hop reasoning, we look into Hotpot-QA \citep{yang2018hotpotqa}  and Musique-Ans \citep{trivedi2022musique2}. The latter includes varying hops and different composition operations, offering a rich test-bed for how well \loci{} can generalize across a diverse range of queries. 

\paragraph{Corpus} \Cref{sec:indexing} describes multiple strategies to index the corpus. Each type of retrieval key needs to be accompanied with its own corpus. Title, passage and sentence keys are derived from the Wikipedia corpus presented in \citealt{bohnet2022attributed}. 
For propositions, we re-use the Factoid-Wiki corpus built by \citealt{chen2023dense}. This is derived from \citealt{bohnet2022attributed} by decomposing passages into smaller, compact propositions using a finetuned Flan-T5-large \citep{wei2021finetuned} model. We drop the titles from Factoid-Wiki and only use the propositions (See \Cref{app:eval_datasets}).

\subsection{Evaluation}
\label{sec:exp:evaluation}
The standard metric for ODQA benchmarks has predominantly been F1 answer match accuracy.
However, language models are prone to hallucinate and F1 stand-alone can be misleading as the answer may not be conditioned on the evidence. Attribution \citep{rashkin2021measuring} helps us trade-off answer accuracy for faithfulness to the evidence.
Thus, we measure two competing metrics: i) end-to-end answer accuracy with F1 and ii) attribution of the answer to evidence using AutoAIS  \citep{bohnet2022attributed}.  
AutoAIS, or AIS for short, is automatically computed by classifying whether the evidence text entails the question and predicted answer pair. We re-use the NLI scorer and formulation from \citealt{bohnet2022attributed} (See details in \Cref{app:eval_datasets}).
The evidence text here is the concatenation of all retrieval keys in the \loci{} output. The unconstrained thoughts are discarded from evaluation.
Only the \textit{top beam output} is considered for evaluation.

\subsection{Models and Inference}
\label{sec:exp:models}
Throughout our experiments, we use off-the-shelf instruction-tuned models in a few-shot setting, without any fine-tuning. We test the instruction-tuned versions of \llmm{} and its larger variant \llml{}~\citep{anil2023palm} based on stacked Transformer architecture. We use 3 example demonstrations in our prompt (\Cref{app:experiment_details}), with different sets of examples for single-hop (NQ) and multi-hop (Hotpot, Musique) datasets. The unconstrained sequences or thoughts are formulated as hint keywords.
Our final setup uses a beam of 10 with constrained decoding (\Cref{sec:constrained}), adaptive beam size (\Cref{sec:adaptive}) and propositions as retrieval keys. Later in \Cref{sec:results}, we ablate these choices. Note that only the \textit{top-beam} result is considered for evaluation.

\subsection{Baselines}
\label{sec:exp:baselines}
We experiment with 3 types of baselines: no retrieval, the standard dense retriever and an iterative retriever suited for multi-hop QA. We opt for baselines that test the out-of-domain system performance, similar to \loci{}.
\paragraph{No Retrieval} We compare to a few-shot unconstrained baseline using the same 3-shot prompt as \loci{} allowing for chain-of-thought reasoning. The setup generates an answer along with hallucinated evidences, not grounded to a corpus. This is a measure of model's memorization capabilities.
\paragraph{Generalized Dense Retriever} For single-hop QA, we compare our approach against the Generalized T5 retriever (GTR-xxl, 11B variant) \citep{ni2021large}. GTR undergoes multi-staged training, first on unsupervised web-mined corpus and then supervised search datasets including NQ. It has been shown to generalize well out-of-domain. 
However, GTR and other conventional dense retrievers provide only retrieved documents, not the answers themselves. To extract answers, we use the \llmm{} model in a few-shot setting (see \Cref{app:experiment_details} for the details).

Since \loci{} generates a single output with a varying number of interleaved documents, direct comparison with dense retrievers that fetch a fixed top-k documents is challenging. We set k to a value equivalent to the mean documents \loci{} fetches for single-hop. When retrieval keys are different, such as passages vs propositions, we approximately match the tokens used by both setups. In our final experiments, we compare against k=1 passage and k=2 propositions for GTR-xxl.

\paragraph{Iterative Retrieval (\mydecomp{})}
For Multi-hop QA, we adopt a popular method where question is decomposed into sub-queries \citep{khot2022decomposed}. At each step, passages are retrieved for a sub-query and fed as input for the next query, until one converges to an answer.
The method has the same surface form as \loci{}, except for the key distinction that each step requires switching between a heterogeneous mix of models. 
In our experiments, we retrieve top-1 document with GTR-xxl and use \llmm{} few-shot for both decomposing the query and generating the final answer (See prompt at \Cref{app:experiment_details}). Max allowed steps is set to 4 where most of the queries converge.

\section{Results}
\label{sec:results}
\begin{table}[ht]
\begin{center}

\begin{tabular}{l|c}
\toprule
Retrieval Key  &  Hits@1 \\
\midrule
Title  & 19.5 \\
Paragraph with Title &  15.5 \\
Paragraph & 19.0 \\
Sentence with Title &  19.1 \\
Sentence &  20.6 \\
Proposition & 33.9 \\
\bottomrule
\end{tabular}

\caption{Comparison of Retrieval Keys on NQ}
\label{tab:search_unit}

\end{center}
\end{table}
\begin{table*}[ht]
\begin{center}

\begin{tabular}{ll|cc|cc|cc}
\toprule
\multirow{2}{*}{Retriever} & \multirow{2}{*}{Answerer} & \multicolumn{2}{c}{NQ} & \multicolumn{2}{c}{Hotpot} & \multicolumn{2}{c}{Musique} \\
& &  F1 & AutoAIS & F1 & AutoAIS & F1 & AutoAIS \\ 
\midrule
\multicolumn{8}{l}{\textit{No Retrieval}} \\
\midrule
\multicolumn{2}{c|}{Unconstrained {\small \llmm{}}} & 41.4 & - & 39.1 & - & 18.3 & - \\
\midrule
\multicolumn{8}{l}{\textit{Dense Retrieval}} \\
\midrule
GTR Passage &  \small{\llmm{}} & 41.9 & 48.7 & 34.9 & 19.6 & 7.2 & 17.9 \\
GTR Proposition &  \small{\llmm{}} & 36.6 & 63.2 & 27.4 & 18.5 & 10.5 & 20.4 \\
Iterative & {\small \llmm{}} & 34.4 & 66.8 & 34.2 &	30.9 & 17.5 & 38.4 \\
\midrule
\multicolumn{6}{l}{\textit{\textit{\loci{}}}} \\
\midrule
\multicolumn{2}{c|}{\small \llmm{}} & 40.2 & 59.2 & 41.0 & 36.5 &  19.1 & 39.6  \\
\multicolumn{2}{c|}{\small \llml{}} & 46.7 & 59.6 & 51.1 & 35.6 &  28.2 & 37.5  \\
\bottomrule
\end{tabular}

\caption{Overall performance comparison for \loci{}. For Dense retrievers, top-k documents are retrieved and fed to the few-shot Answerer, where k=1 for GTR passage, k=2 for GTR propositions. For Iterative retrieval upto 4 documents are retrieved with k=1 at each step.}
\label{tab:results}

\end{center}
\end{table*}

\begin{table}[ht]
\begin{center}

\begin{tabular}{l|c|c}
\toprule
Beam  & F1  & AutoAIS \\
\midrule
1 & 19.3 &	26.1 \\
5 & 35.8 &	58.7\\	
10 & 40.2 &	59.2 \\
\bottomrule
\end{tabular}

\caption{Effect of Beam size on NQ with \llmm{}.}
\label{tab:beam}

\end{center}
\end{table}

\begin{table}[ht]
\begin{center}

\begin{tabular}{c|c|c|c|c|c}
\toprule
{\small Unconst.} & {\small Adaptive}  & \multicolumn{2}{c}{NQ} & \multicolumn{2}{c}{Hotpot} \\
{\small Keywords} &  {\small Beam}  & F1 & AIS & F1 & AIS \\
\midrule
\text{\sffamily X} & \text{\sffamily X} & 37.9 & 57.5 & 39.2 & 33.9 \\
 \checkmark & \text{\sffamily X} & 36.9 & 51.5 & 38.4 & 32.3 \\
 \checkmark & \checkmark & 40.2 & 59.2 & 41.0 & 36.5 \\
\bottomrule
\end{tabular}

\caption{Interleaving unconstrained keywords and retrieval keys with Adaptive beam. Greedily decoding Unconstrained sub-sequences allows constrained retrievals to make the most of the beam search.
}
\label{tab:blend}
\end{center}
\end{table}

In the following sections, we investigate the key building blocks of \loci{}: i) indexing strategies (\Cref{sec:indexing}) amenable to auto-regressive decoding ii) effect of beam decoding (\Cref{sec:constrained}) iii) suitable mechanisms to interleave thoughts and retrieval keys (\Cref{sec:adaptive}). Finally, we compare \loci{} against conventional retrieval systems. We also draw a detailed analysis of wins and losses to fathom the strengths and pitfalls of the system.

\subsection{\loci{} building blocks}
\paragraph{Retrieval Keys}
\label{sec:results:retrieval_keys}
We explore the following retrieval key candidates as detailed in \Cref{sec:indexing}:
a) \textit{Title}: Wikipedia page and section titles, ranking paragraphs within the section using TF-IDF scores.
b) \textit{Paragraph with Title}: Decodes the page title, section title, and full paragraph.
c) \textit{Paragraph}: Decodes the paragraph only.
d) \textit{Sentence}: Uses individual sentences.
e) \textit{Proposition}: Uses atomic information units derived from paragraphs.
Table~\ref{tab:search_unit} shows that among the retrieval keys explored, the propositional index is best aligned with our decoding search strategy, perhaps its compact nature is most suited for autoregressive decoding. An in-depth analysis of retrieval keys is provided in \Cref{app:index_units}. 
In the following experiments, we use proposition as our retrieval key.

\paragraph{Effect of Beam size}
\Cref{tab:beam} shows how greedy decoding can get stuck with poor retrieval keys. A larger beam  enables better search space exploration, albeit with diminishing returns. In our final experiments, we use a beam of 10.

\paragraph{Interleaving with Adaptive Beam}
\label{sec:results:blend}
\Cref{tab:blend} shows the impact of interleaving thoughts with retrieval keys. 
First, we note that an adaptive beam is crucial for interleaving unconstrained and constrained sequences. Without an adaptive beam, minor irrelevant variations in unconstrained thoughts can consume and overwhelm the available space in the beam. By greedily decoding unconstrained sequences, the beam space is preserved for backtracking during document search.
Once we have an adaptive beam in place, the insertion of keywords enhances both answer and retrieval performance, reminiscent of chain-of-thought technique to enable better retrieval.

\subsection{Overall Results}
\label{sec:results:overall}
\Cref{tab:results} shows the overall performance of \loci{} across various datasets. First, we compare with the no-retrieval baseline and observe that jointly retrieving and answering does not negatively impact model's answering capabilities.
For single-hop tasks, \loci{} competes well with dense retrievers, offering higher answer accuracy at the expense of attribution. In multi-hop QA, \loci{} excels, outperforming iterative baselines by +15 F1 points on Hotpot and +11 on Musique, with comparable or better attribution. The increase in answer accuracy with the larger \llml{} model suggests improved performance with larger model sizes. Notably, \loci{} achieves these results with a single inference pass, unlike the Iterative baseline, which requires a model call at each sub-query step.

\subsection{Qualitative analysis}
\label{sec:results:qualitative}
We inspect 50 win and loss examples each to analyze the strength and weaknesses of the system.
\paragraph{Wins} Several properties distinguish \loci{} from dense retrievers: a) \loci{} allows large language models to utilize their parametric knowledge for retrieval. Since the search operation in \loci{} is independent of corpus size, it can employ much larger models at query time. b) The inherent alignment of instruction-tuned models enables them to retrieve contextually relevant passages, whereas dense retrievers may sometimes latch onto keywords. c) The interleaved thoughts guide the model toward more accurate retrievals. \Cref{tab:wins} demonstrates these scenarios for single-hop retrievals and \Cref{tab:win_multihop} for multi-hop retrievals.

\paragraph{Can the model retrieve what it doesn't know?} A language model may hold stale or incorrect information. However, \loci{} can often override model's pre-existing knowledge and generate correct answers by constraining on the corpus (\Cref{app:qualitative})

\begin{table}[ht]
\begin{center}

\begin{tabular}{l|c}
\toprule
Failure mode  & Queries(\%) \\
\midrule
Index Failure & 40\% \\
Search Failure & 52\%  \\
Attribution Failure & 8\% \\
\bottomrule
\end{tabular}

\caption{Loss categories for \loci{} on Hotpot-QA}
\label{tab:loss_category}

\end{center}
\end{table}
\paragraph{Losses} We inspect 50 failed queries and categorize the losses (\Cref{tab:loss_category}) as follows: a) Index failure: the proposition is absent from the index or not decontextualized. b) Search failure: Proposition exists in the index, but could not be generated c) Attribution failure: The answer is partially attributed, with LLM hallucinating based on partial evidence. (see \Cref{app:qualitative} for examples)

\section{Conclusion}
Retrieval has so far been alienated from the rapid progress in instruction tuning. This work makes the following contribution: 
i) an approach that can seamlessly integrate retrieval with generation. ii) a thorough investigation of indexing and search strategies that enable such an approach to be effective. iii) proof-of-concept of the capabilities of such a system on a variety of QA tasks.
We hope the ideas introduced in this work fuel progress in aligning retrieval to generation and simplifying it.
\section{Limitations}
First we note the limitations in our experimental setup. All our experiments are based on Wikipedia, a corpus heavily seen during pre-training. This work does not analyze how \loci{} fares on corpora unseen during pre-training. Furthermore, we only examine a handful of factoid question-answering tasks due to the lack of objective evaluations. Performance on tasks such as long-form QA is deferred for future work.
There are also certain inherent limitations with \loci{}. It forces verbatim emission of corpus text, which might be an overkill for tasks where a similarity-based metric is sufficient. \loci{} lacks the ability to retrieve dozens of documents, a necessity for certain summarization tasks. For long documents with diffused information, rewriting into propositions adds complexity and can be cumbersome.
Lastly, while \loci{}'s search operation is independent of corpus size, the use of beam search and communication between the FM-index and Transformer model can slow down inference.
\section{Ethical Considerations}
All artifacts used in this paper, including models, datasets, and baselines, are under permissive licenses and publicly available. We have attempted to provide detailed information to facilitate the reproduction of our results. 

Our findings are based on English-language data from Wikipedia, and we have not tested the generalizability of our claims to other languages or domains.

Lastly, the datasets used in this work are not expected to contain any offensive content. However, it is important to note that Large Language Models (LLMs) can exhibit biases related to gender, race, and region, and are also prone to hallucination. Although \loci{} aims to ground its generation in an external corpus, some biases may still be present.

\bibliography{anthology,custom}
\bibliographystyle{acl_natbib}

\clearpage
\appendix
\section{Appendix}

\begin{table}[ht]
\begin{center}

\begin{tabular}{l|c|c|c}
\toprule
Dataset  & Split & Queries & Hops \\
\midrule
Open-NQ & Test & 3610 & 1\\
Hotpot  & Dev & 7405 & 2 \\
MuSiQue-Ans & Dev & 2412 & 2-4 \\
\bottomrule
\end{tabular}

\caption{ODQA Datasets used in our experiments}
\label{tab:dataset}

\end{center}
\end{table}

\begin{table}[ht]
\begin{center}

\begin{tabular}{l|c|c}
\toprule
Corpus  & Docs &  Avg Words \\
\midrule
Passage & 40M & 58.5 \\
Sentence & 114M & 21.0 \\
Propositions & 256M & 11.0 \\
\bottomrule
\end{tabular}
\caption{Retrieval Corpora used in our experiments}
\label{tab:corpus}

\end{center}
\end{table}

\subsection{Experiment Details}
\label{app:experiment_details}
\begin{table*}[ht]
\begin{center}

\begin{tabular}{|p{\linewidth}|}
\toprule
For given input query, write 1-3 passages to answer the query.
Write a hint keyword and a passage contained within << and >>.
A passage must be a complete sentence and not a phrase. It must contain complete context for answering the query and should not begin with it, he, they etc.
Do not repeat any passages. Aim for new keywords. \\
\\

question: The football manager who recruited Cristiano Ronaldo managed Manchester United during what timeframe? \\
passage: keyword: Cristiano Ronaldo's recruiting manager << Alex Ferguson recruited Cristiano Ronaldo >> keyword: Sir Alex Ferguson's tenure at Manchester United << Sir Alex Ferguson managed Manchester United from 1986 to 2013. >> \\
answer: 1986 to 2013 \\
\\

question: Were Eatza Pizza and Your Pie founded in the same state? \\
passage: keyword: Eatza Pizza founded in state << Eatza Pizza was founded in Arizona >> keyword: Your Pie founded in state << Your Pie was founded in Athens, Georgia >> \\
answer: no \\
\\
question: In which stadium do the teams owned by Myra Kraft's husband play? \\
passage: keyword: Myra Kraft's husband << Robert Kraft's wife is Myra Kraft. >> keyword: Robert Kraft's team << Robert Kraft is the owner of the New England Patriots. >> keyword: New England Patriots stadium << Gillette Stadium is the home of the New England Patriots. >> \\
answer: Gillette Stadium \\
\\
question: <question> \\
passage: \\
\bottomrule
\end{tabular}
\caption{Few-shot prompt used in \loci{} for multi-hop QA}
\label{prompt:proposition}
\end{center}
\end{table*}
\begin{table*}[ht]
\begin{center}

\begin{tabular}{|p{\linewidth}|}
\toprule
For given input query, write 1-3 passages to answer the query.
Write a hint keyword and a passage contained within << and >>.
A passage must be a complete sentence and not a phrase. It must contain complete context for answering the query and should not begin with it, he, they etc.
Do not repeat any passages. Aim for new keywords. \\
\\
question: who is the owner of phoenix mall pune? \\
passage: keyword: Phoenix Market City owner << Phoenix Market City is developed by Phoenix Mills Limited. >> \\
answer: Phoenix Mills Limited \\
\\
question: what brings in more money nba or nfl? \\
passage: keyword: NFL revenues << NFL revenues are well over \$10 billion per season. >> keyword: NBA revenue << NBA amasses about \$6 billion annually. >> \\
answer: NFL \\
\\
question: when was the french national anthem adopted? \\
passage: keyword: French national anthem << La Marseillaise  became the national anthem of France. >> keyword: La Marseillaise adoption << La Marseillaise was adopted by France in 1795. >> \\
answer: 1795 \\
\\
question: {question} \\
passage: \\
\bottomrule
\end{tabular}
\caption{Few-shot prompt used in \loci{} for single-hop QA}
\label{prompt:single_hop}
\end{center}
\end{table*}
\paragraph{In-context prompts}
We use 2 different sets of few-shot demonstration for single-hop (NQ) and multi-hop (Hotpot, Musique) datasets displayed in \Cref{prompt:proposition} and \Cref{prompt:single_hop} respectively. Both prompts carry the same instruction, but the multi-hop variants provides demonstrations with multiple evidence passages.

\begin{figure*}[t]
\includegraphics[width=\textwidth]{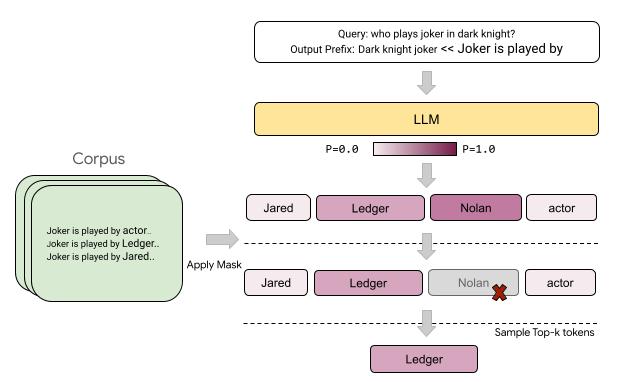}
\caption{Illustration of the constrained decoding process. Given prefix, "Joker is played by", the continuation "Nolan" is not found in the corpus and therefore masked out.}

\label{fig:approach}
\end{figure*}
\paragraph{Computing constraints}
An example of constrained decoding is illustrated in \Cref{fig:approach}.

\begin{table*}[ht]
\begin{center}

\begin{tabular}{|p{\linewidth}|}
\toprule
Answer the 'question' only based on the given 'passage'.
If the 'passage' lacks context or is not relevant, say 'Cannot answer' else say generate a short answer.
Do not answer the query from outside the scope of the passage. \\
\\

question: what brings in more money nba or nfl? \\
passage: NFL revenues are well over \$10 billion per season.
NBA amasses about \$6 billion annually. \\
answer: NFL \\
\\
question: when did they put warnings on cigarette packs \\
passage: Tobacco packaging 1978's warning was not removed, so now every cigarette pack contains both warnings (one on each lateral). \\
answer: Cannot Answer \\
\\
question: when was the french national anthem adopted? \\
passage: La Marseillaise  became the national anthem of France. 
La Marseillaise was adopted by France in 1795. \\
answer: 1795 \\
\\
question: {question} \\
passage: {passage} \\
answer: 
\\
\bottomrule
\end{tabular}
\caption{Few-shot prompt for extracting answer from propositions}
\label{prompt:reader}
\end{center}
\end{table*}
\begin{table*}[ht]
\begin{center}
\begin{tabular}{|p{0.9\linewidth}|}
\toprule
You are given a multi-hop `question`. Decompose it into simple single-hop query, passage. And finally write the overall answer. \\
\\
question: In what country was Lost Gravity manufactured? \\
query: Who manufactured The Lost Gravity (roller coaster)? \\
passage: Lost Gravity is a steel roller coaster at Walibi Holland manufactured by Mack Rides. \\
query: Mack Rides is from which country? \\
passage: Mack Rides is based in Germany. \\
answer: Germany \\
\\
question: Do James Cameron and Christopher Nolan share their profession? \\
query: What is the profession of James Cameron? \\
passage: James Cameron is a Director. \\
query: What is the profession of Christopher Nolan? \\
passage: Christopher Nolan is a Director. \\
answer: Yes \\
\\
question: The actor that stars as Joe Proctor on the series "Power" also played a character on "Entourage" that has what last name? \\
query: Who is the actor that stars as Joe Proctor on the series "Power"? \\
passage: Joe Proctor on the series "Power" was potrayed by Jerry Ferrara. \\
query: Jerry Ferrara played a character on Entourage named what? \\
passage: Jerry Ferrara played the character of Assante on Entourage. \\
answer: Assante \\
\\
question: <question> \\
<sub-query steps so far> \\
\bottomrule
\end{tabular}
\caption{Few-shot prompt for Iterative baseline}
\label{prompt:iterative}
\end{center}
\end{table*}
\paragraph{Baselines}
For the dense-retriever baseline, answers are extracted from retrieved passages with an external reader. We use \llmm{} with a few-shot prompt (\Cref{prompt:reader}).

For iterative retrieval baseline, we use \llmm{} for both query decomposition and answering. At each step, the model can choose to generate a sub-query or the final answer. The unified prompt is provided at \Cref{prompt:iterative}.
 

\subsection{Evaluation}
\label{app:eval_datasets}
\paragraph{Datasets} We use Musique-Ans \citep{trivedi2022musique2} subset of Musique which consists of answerable queries.
Details of query sets evaluated can be found in \Cref{tab:dataset}.
To make retrieval challenging, we use the full Wikipedia corpus for retrieval (\Cref{tab:corpus}). This is different from the typical Hotpot and Musique  setting which use the first Wikipedia paragraph (5M documents) and documents associated with query-set (1.3M) respectively.

\paragraph{Metrics}
AutoAIS is an automated way of measuring AIS (Attributable to Identified Source) \citep{rashkin2021measuring}. AutoAIS formulates evaluation as a Natural Language Inference task that asks a model whether the
question and answer are entailed by the provided evidence. We re-use a T5-11B checkpoint finetuned on a collection of NLI tasks from \citep{bohnet2022attributed}. 
Question answering is formulated into NLI task as follows: 

\texttt{\small hypothesis: <retrieved evidence1> <retrieved evidence2> ... premise: The answer to the question '<question>' is '<predicted answer>'
}
The NLI scorer provides a calibrated score between 0 (not attributed) to 1 (completely attributed) which is averaged over the dataset to get the final AutoAIS score.

\begin{table*}[ht]
\begin{center}
\small
\begin{tabular}{p{0.35\textwidth}|p{0.35\textwidth}|p{0.2\textwidth}}
\toprule
Query & Retrievals & Comment \\
\midrule
\multicolumn{3}{l}{\normalsize Index failure} \\
\midrule
how many episodes of touching evil are there &  A total of 35 episodes were produced.  & Proposition lacks context \\
\midrule
who is the coach for the ottawa senators & D. J. Smith is the head coach of the Ottawa Senators.  & Incorrect Proposition generated \\
\midrule
\multicolumn{3}{l}{\normalsize Search failure} \\
\midrule
what age do you need to be to buy a bb gun & 18 years of age or older.  & partial phrase decoded \\
\midrule
how many seasons of the bastard executioner are there & 
The Bastard Executioner is an American historical fiction drama television series.
\newline 
The Bastard Executioner is an American historical fiction drama television series. & repeated retrieval \\
\midrule
who plays gram on the young and the restless & 
The Young and the Restless is an American television soap opera.
\newline
The Young and the Restless was first broadcast on March 26, 1973. & irrelevant \\
\bottomrule
\end{tabular}

\caption{Example losses in \loci{}}
\label{tab:losses}

\end{center}
\end{table*}

\begin{table*}[ht]
\begin{center}
\small
\begin{tabular}{p{0.45\textwidth}|p{0.45\textwidth}}
\toprule
Unconstrained Generation & Constrained Generation  \\
\midrule
\multicolumn{2}{c}{\textit{Q: who was the actor that played ben stone on law and order}} \\
Ben Stone was played by actor Jerry Orbach. & Ben Stone was played by Michael Moriarty.\\
\midrule
\multicolumn{2}{c}{\textit{Q: how many pieces in a terry's chocolate orange}} \\
Terry's Chocolate Orange is made with 32 segments & Terry's Chocolate Orange is divided into 20 segments \\
\midrule
\multicolumn{2}{c}{\textit{Q: who sings the song only in my dreams}} \\
The song "Only in My Dreams" is sung by the band Air Supply. & Only in My Dreams is the debut single by Debbie Gibson. \\
\bottomrule
\end{tabular}
\caption{Unconstrained vs Constrained generation. Examples where unconstrained LLM emits incorrect answer but constraining on the corpus helps \loci{} override this pre-existing knowledge to obtain the correct answer}
\label{tab:override}
\end{center}
\end{table*}


\subsection{Extended Qualitative Analysis}
\label{app:qualitative}
We provide examples for loss categories defined in \Cref{sec:results} in \Cref{tab:losses}.
\Cref{tab:override} showcases a few selected examples where the unconstrained model emits incorrect answer, but constraining on the corpus guides it towards correct answer.

\subsection{Index representation qualitative analysis}
\label{app:index_units}
In this section we analyse retrievals from a few select examples from the OpenNQ development dataset. We compare retrievals using different document representations in the index, including: \textbf{title + tf-idf}, \textbf{passage}, \textbf{sentence} and \textbf{proposition} (see Section~\ref{sec:indexing}).

\begin{small}

\vspace{4pt}\noindent\hrule height 0.8pt\vspace{4pt}

\noindent
\textbf{Question} who holds the world record for the most world records \\
\textbf{Gold Answer} Ashrita Furman
\begin{itemize}[noitemsep,nolistsep,leftmargin=*]
\item \textbf{title + tf-idf}: List of world records in track cycling. World records in the sport of track cycling are ratified by the Union Cycliste Internationale (UCI).
\item \textbf{Passage}: According to the Guinness Book of World Records, the record for the largest event was set in 2009.
\item \textbf{Sentence}: As of January 2021, he has 1.7 million followers on Twitter
\item \textbf{Proposition}: \textit{\textbf{Ashrita Furman}} holds the Guinness world record for the most Guinness world records.
\item \textbf{Source text for proposition}: Ashrita Furman (born Keith Furman, September 16, 1954) is a Guinness World Records record-breaker. As of 2017, Furman has set more than 600 official Guinness Records and currently holds 531 records, thus holding the Guinness world record for the most Guinness world records.
\end{itemize}
\textbf{Analysis} In this example, only the propositional-based retrieval was able to retrieve a correct answer. Part of the reason why passage or sentence representation is hard to retrieve with auto-regressive decoding is that the main evidence for this answer in the "source text for proposition" comes at the end of a complex sentence ("[...] thus holding the Guiness world record for the most Guiness world records").

\vspace{4pt}\noindent\hrule height 0.8pt\vspace{4pt}

\noindent
\textbf{Question} who has the most number one single hits \\
\textbf{Gold Answer} The Beatles
\begin{itemize}[noitemsep,nolistsep,leftmargin=*]
\item \textbf{title + tf-idf}: List of Billboard Hot 100 number-one singles of the 1990s. This article is about the American Billboard Hot 100 chart held during the 1990s.
\item \textbf{Passage}:  According to the Recording Industry Association of America (RIAA), the album has sold more than 500,000 copies in the United States,
\item \textbf{Sentence}: As of 2021, \textbf{\textit{The Beatles}} remain the act with the most Christmas number-one singles in the United Kingdom.
\item \textbf{Proposition}: \textit{\textbf{The Beatles}} hold the record for most number-one hits on the Billboard Hot 100.
\item \textbf{Source text for proposition}: The Beatles, Awards and achievements. As of 2017, they hold the record for most number-one hits on the Billboard Hot 100, with twenty.
\end{itemize}
\textbf{Analysis} The correct answer is retrieved in this example using both sentence and proposition based representations. However, the sentence representation here does not provide proper justification for the answer. Note that the propositional representation correctly decontextualizes the pronoun in "[...] they hold the record [...]" to "The Beatles hold the record [...]" making the retrieval easier using constrained decoding.

\vspace{4pt}\noindent\hrule height 0.8pt\vspace{4pt}

\noindent
\textbf{Question} how many episodes of sabrina the teenage witch are there \\
\textbf{Gold Answer} 163
\begin{itemize}[noitemsep,nolistsep,leftmargin=*]
\item \textbf{title + tf-idf}: Sabrina the Teenage Witch (1996 TV series). The first four seasons aired on ABC from September 27, 1996 to May 5, 2000. The final three seasons ran on The WB from September 22, 2000 to April 24, 2003.
\item \textbf{Passage}: Sabrina the Teenage Witch is an American television sitcom created by Nell Scovell, based on the Archie Comics series of the same name.
\item \textbf{Sentence}: Sabrina the Teenage Witch is an American television sitcom created by Nell Scovell, based on the Archie Comics series of the same name.
\item \textbf{Proposition}: Sabrina the Teenage Witch had \textbf{\textit{163}} episodes.
\item \textbf{Source text for proposition}: This is an episode list for Sabrina the Teenage Witch, an American sitcom that debuted on ABC in 1996. From Season 5, the program was aired on The WB. The series ran for seven seasons totaling 163 episodes.
\end{itemize}
\textbf{Analysis} All retrievals using non-propositional representations select part of the main article for "Sabrina the Teenage Witch". This article, however, does not contain the answer to the question. In the propositional case, there is a straightforward proposition that is constructed from a passage from the "List of Sabrina the Teenage Witch episodes". Note that the source passage contains a reference that becomes ambiguous out-of-context ("The series" is decontextualized to "Sabrina the Teenage Witch" in the proposition).

\vspace{4pt}\noindent\hrule height 0.8pt\vspace{4pt}

\noindent
\textbf{Question} what is dj's boyfriends name on full house \\
\textbf{Gold Answers }Steve Hale, Steven "Steve" Hale, rich kid Nelson, or Viper
\begin{itemize}[noitemsep,nolistsep,leftmargin=*]
\item \textbf{title + tf-idf}: Full House (season 8). The eighth and final season of the ABC sitcom Full House originally aired between September 27, 1994 and May 23, 1995.
\item \textbf{Passage}:  Full House (1987–1995) and its Netflix sequel Fuller House.
\item \textbf{Sentence}: In the 1990s, she appeared in the films Blues Brothers 2000
\item \textbf{Proposition}: \textbf{\textit{Steve Hale}} was D.J.'s boyfriend in seasons six and seven.
\item \textbf{Source text for proposition}: Full House, Production, Casting. As babies, the children were played by Daniel and Kevin Renteria, and in season six, the roles of the twins were succeeded by Blake and Dylan Tuomy-Wilhoit. The last main character added was Steve Hale, who was D.J. 's boyfriend in seasons six and seven. He was played by Scott Weinger.
\end{itemize}
\textbf{Analysis} The source sentence with the correct answer presents a challenge for auto-regressive decoding since the sentence prefix focuses on an aspect unrelated to the question ("The last main character added [...]"). With propositionalization, the sentence structure becomes aligned with the question, but requires that the model already knows the answer to the question, given that the first entity in the sentece is the answer.

\vspace{4pt}\noindent\hrule height 0.8pt\vspace{4pt}

\noindent
\textbf{Question} who is the girl in green day 21 guns \\
\textbf{Gold Answer} Lisa Stelly
\begin{itemize}[noitemsep,nolistsep,leftmargin=*]
\item \textbf{title + tf-idf}: Boulevard of Broken Dreams (Green Day song), Music video. The video won six awards at the MTV Video Music Awards in 2005, most notably for Video of the Year. It also won Best Group Video, Best Rock Video, Best Direction, Best Editing, and Best Cinematography.
\item \textbf{Passage}: "21 Guns" is a song by American rock band Green Day. It was released as the second single from their eighth studio album, 21st Century Breakdown (2009), and serves as the sixteenth track from the album. The single was released through Reprise Records on May 25, 2009 as a digital download and July 14, 2009 as a CD single.
\item \textbf{Sentence}: "21 Guns" is a song by American rock band Green Day.
\item \textbf{Proposition}: The girl in the music video is Teresa Lourenco.
\item \textbf{Source text for proposition}: The music video for \"Again\" features Kravitz with his girlfriend in his apartment (Gershon), whom he does not seem to be interested in. Similar to the song's lyrical content, he meets a girl (Teresa Lourenco), who works as a waitress in a restaurant/diner.
\end{itemize}
\textbf{Analysis} In this case, all retrievals fail to retrieve the correct answer. In the case of the proposition-based representation, the model decodes a proposition where the subject is an ambiguous reference ("The girl") which has not been properly decontextualized (the source passage above makes it clear that the reference is not related to the question). Interestingly, the source passage with the correct answer requires an inferential step and its proposition representations are been decontextualized properly. \textbf{Source text with correct answer}: \textit{21 Guns (song), Music video. The video takes place with the band and the album's two protagonists Christian (Josh Boswell) and Gloria (Lisa Stelly) taking refuge in a white room after robbing a bank.}.\\
\textbf{Relevant generated propositions}: 
\begin{itemize}[noitemsep,nolistsep,leftmargin=*]
\item The video takes place with the band and the album's two protagonists Christian and Gloria.
\item Gloria is played by Lisa Stelly.
\end{itemize}
To properly retrieve this passage using proposition-based representation we would need to properly disambiguate "The video" to "21 guns" and perform inference over these two propositions. Alternatively, proposition generation could generate more complex propositions containing both pieces of information, such as: \textbf{The "21 Guns" video takes place with the protagonist Gloria, played by Lisa Stelly}.

\vspace{4pt}\noindent\hrule height 0.8pt\vspace{4pt}

\noindent
\textbf{Question} how many seasons of vampire diaries r there  \\
\textbf{Gold Answer} eight, or 8
\begin{itemize}[noitemsep,nolistsep,leftmargin=*]
\item \textbf{title + tf-idf}: The Vampire Diaries. The Vampire Diaries is an American supernatural teen drama television series developed by Kevin Williamson and Julie Plec, based on the book series of the same name written by L. J. Smith. The series premiered on The CW on September 10, 2009, and concluded on March 10, 2017, having aired 171 episodes over \textbf{\textit{eight}} seasons.
\item \textbf{Passage}:  The Vampire Diaries is an American supernatural teen drama television series developed by Kevin Williamson and Julie Plec, based on the book series of the same name written by L. J. Smith. The series premiered on The CW on September 10, 2009, and concluded on March 10, 2017, having aired 171 episodes over \textit{\textbf{eight}} seasons.
\item \textbf{Sentence}: The series premiered on The CW on September 10, 2009, and concluded on March 10, 2017, having aired 171 episodes over \textbf{\textit{eight}} seasons.
\item \textbf{Proposition}: The Vampire Diaries is an American supernatural drama television series.
\item \textbf{Source text for proposition}: The Vampire Diaries is an American supernatural drama television series that premiered on The CW on September 10, 2009, and concluded on March 10, 2017 after airing eight seasons.
\end{itemize}
\textbf{Analysis} In this case only the proposition-based representation retrieval is incorrect. We believe the retrieval fails here due to improper decontextualization of the correct answer passage. The sentence with the correct answer includes the proposition: \textit{The series aired 171 episodes over eight seasons.}. Making it difficult for the model to 

\vspace{4pt}\noindent\hrule height 0.8pt\vspace{4pt}


\end{small}

\subsection{Computations involved}
Evaluating the precise compute cost for \loci{} depends on the specific implementations of the decoding algorithm, but we can sketch the key operations involved in retrieval: indexing and search.
Indexing depends on the number of items in the corpus \( |D| \). We use a model of size \( \mathcal{M} \) to rewrite each passage (average length \( |p| \)) into propositions. The overall indexing cost is proportional to \( O(D\mathcal{M}p^2) \), similar in magnitude to the cost for encoding the corpus in dense retrieval, differing only by a constant factor. Note that our experiments use a T5-large backbone (770M) for \loci{} much smaller than T5-xxl (11B) used in the dense baselines.

Now let's look at the search operation. At each auto-regressive step, besides standard decoding, the only additional operation is computing FM-index constraints, which consumes CPU resources. However, while the index is efficient, communication between the index on the host and the Transformer model on the GPU/TPU adds latency to the decoding step. In contrast, RAG systems retrieve documents from index using nearest neighbor scan in a single go. But even there, the documents need to encoded as input to the language model.





\end{document}